\documentclass[review]{elsarticle}

\usepackage{caption}
\usepackage{subcaption}
\usepackage{algorithmic}
\usepackage{algorithm}
\usepackage{graphicx}
\usepackage{textcomp}
\usepackage{xcolor}
\usepackage{bbm}
\usepackage{algorithm}
\usepackage{caption}
\usepackage{subcaption}
\usepackage{enumitem}
\usepackage{booktabs}
\usepackage{amsfonts}
\usepackage{amsmath}
\usepackage{hyperref}
\usepackage{multirow}
\usepackage{mathtools}

\usepackage{booktabs}
\usepackage{hyperref}
\usepackage{subcaption}
\usepackage{adjustbox}
\usepackage{multirow}

\DeclareMathOperator*{\argmin}{arg\,min}
\newcommand\norm[1]{\left\lVert#1\right\rVert}
\definecolor{todocolor}{rgb}{0.9,0.1,0.1}
\definecolor{lcolor}{rgb}{0.7,0.7,0.3}
\definecolor{qcolor}{rgb}{0,0,255}

\DeclareSymbolFont{matha}{OML}{txmi}{m}{it}
\DeclareMathSymbol{\varv}{\mathord}{matha}{118}

\journal{Journal of \LaTeX\ Templates}

\newtheorem{definition}{Definition}[section]

\newproof{Example}{Example}
\newproof{Method}{Method}
\newproof{Exercise}{Exercise}
\newproof{proof}{Proof}

\begin{document}

\begin{frontmatter}

\title{Causality-based Counterfactual Explanation for Classification Models}


\author[mymainaddress]{Tri Dung Duong} 
\ead{TriDung.Duong@student.uts.edu.au}
\address[mymainaddress]{Faculty of Engineering and Information Technology, University of Technology Sydney, \\NSW, Australia}

\author[mysecondaryaddress]{Qian Li \corref{mycorrespondingauthor}} 
\ead{qli@curtin.edu.au}
\address[mysecondaryaddress]{School of Electrical Engineering, Computing and Mathematical Sciences,\\ Curtin University, WA, Australia}

\author[mymainaddress]{Guandong Xu\corref{mycorrespondingauthor}}
\cortext[mycorrespondingauthor]{Corresponding author}
\ead{Guandong.Xu@uts.edu.au}





\begin{abstract}
Counterfactual explanation is one branch of interpretable machine learning that produces a perturbation sample to change the model's original decision. The generated samples can act as a recommendation for end-users to achieve their desired outputs. Most of the current counterfactual explanation approaches are the gradient-based method, which can only optimize the differentiable loss functions with continuous variables.
Accordingly, the gradient-free methods are proposed to handle the categorical variables, which however have several major limitations: 1) causal relationships among features are typically ignored when generating the counterfactuals, possibly resulting in impractical guidelines for decision-makers; 2) the counterfactual explanation algorithm requires a great deal of effort into parameter tuning for dertermining the optimal weight for each loss functions which must be conducted repeatedly for different datasets and settings. In this work, to address the above limitations, we propose a prototype-based counterfactual explanation framework (ProCE). ProCE is capable of preserving the causal relationship underlying the features of the counterfactual data. In addition, we design a novel gradient-free optimization based on the multi-objective genetic algorithm that generates the counterfactual explanations for the mixed-type of continuous and categorical features. Numerical experiments demonstrate that our method compares favorably with state-of-the-art methods and therefore is applicable to existing prediction models. All the source codes and data are available at \url{https://github.com/tridungduong16/multiobj-scm-cf}.



\end{abstract}

\begin{keyword}
counterfactual explanation, interpretable machine learning, structural causal model
\end{keyword}

\end{frontmatter}

\section{Introduction}



Machine learning (ML) is increasingly recognized as an effective approach for large-scale automated decisions in several domains. However, when an ML model is deployed in critical decision-making scenarios such as criminal justice \cite{zavrvsnik2019algorithmic,kaur2020interpreting} or credit assessment \cite{galindo2000credit}, many people are skeptical about its accountability and reliability. Hence, interpretablity is vital to make machine learning models transparent and understandable by humans. Recent years witness an increasing number of studies that have explored ML mechanisms under the causal perspective \cite{schwab2019cxplain,williams2016axis,zhao2021causal}. Among these studies, counterfactual explanation (CE) is the prominent example-based method that focuses on generating counterfactual samples for interpreting model decisions.
For example, 
consider a customer \texttt{A} whose loan application has been rejected by the ML model of a bank.
Counterfactual explanations can generate a ``what-if'' scenario of this person, e.g., ``your loan would have been approved if your income was \$51,000 more''.
Namely, the goal of counterfactual explanation is to generate perturbations of an input that leads to a different outcome from the ML model.
By allowing users to explore such ``what-if'' scenarios, counterfactual examples are interpretable and are easily understandable by humans.

Despite recent interests in counterfactual explanations, existing methods suffer three limitations. First, the counterfactual methods neglect the causal relationship among features, leading to the infeasible counterfactual samples for decision makers \cite{ustun2019actionable,poyiadzi2020face}. 
In fact, a counterfactual sample is considered as feasible if the changes satisfy conditions restricted by the causal relations. 
For example, since education causes the choice of the occupation, changing the occupation without changing the education is infeasible for the loan applicant in the real world.
Namely, the generated counterfactuals need to preserve the causal relations between features in order to be realistic and actionable.  
Second, on the algorithm level,
most counterfactual methods use the gradient-free optimization algorithm to deal with various data and model types \cite{sharma2020certifai,poyiadzi2020face,dhurandhar2019model,grath2018interpretable,lash2017generalized}. 
These gradient-free optimizations rely on the heuristic search, which however suffers from inefficiency due to the large heuristic search space. In addition, optimizing the trade-off among different loss terms in the objective function is difficult, which often leads to sub-optimal counterfactual samples \cite{mahajan2019preserving,mothilal2020explaining,grath2018interpretable}.

To address the above limitations, we propose a prototype-based counterfactual explanation framework (ProCE) in this paper. ProCE is a model-agnostic method and is capable of explaining the classification in the mixed feature space. It should be emphasized that the proposed method focuses on maintaining the causal relationships among the features in dataset instead of the causal relationship between features and target variable \cite{fernandez2020explaining}. Overall, our contributions are summarized as follows:




\begin{itemize}
    \item By integrate causal discovery framework and causal loss function, our proposed method can produce the counterfactual samples that satisfy the causal constraints among features. 
    
    

    \item We utilize the auto-encoder model and class prototype to guide the search progress and speed up the searching speed of counterfactual samples.
    \item We design a novel multi-objective optimization that can find the optimal trade-off between the objectives while maintaining diversity in counterfactual explanations' feature space.
    
\end{itemize}




\section{Background}
\subsection{Preliminary}

Throughout the paper, lower-cased letters $x$ and $\boldsymbol{x}$ denote the deterministic scalars and vectors, respectively. We consider a dataset $\mathcal{D} = \{\boldsymbol{x}_i, c_i\}^n_{i=1}$ consisting of $n$ instances, where $\boldsymbol{x}_i \in \mathcal{X}$ is
a sample, $c_i \in \mathcal{C} = \{0, 1\}$ is the class of
individuals $\boldsymbol{x}_i$, and $\boldsymbol{x}_i^j$ is
the $j$-th feature of $\boldsymbol{x}_i$. Also, we consider a classifier $\mathcal{H}: \mathcal{X} \rightarrow \mathcal{Y}$ 
 that has the input of feature space $\mathcal{X}$ and the output as $\mathcal{Y} = \{0, 1\}$. We denote $Q_{\phi}(.)$ as an encoder model parameterized by $\phi$. Finally, $\text{proto}_*(\boldsymbol{x})$ and $\mathcal{K}(\boldsymbol{x})$ are the prototype and the set of $K$-nearest instances of an instance $\boldsymbol{x}$, respectively.



\begin{definition}[Counterfactual Explanation]

With the original sample $\boldsymbol{x}_\text{org} \in \mathcal{X}$, and original prediction $y_\text{org} \in \mathcal{Y}$, the counterfactual explanation aims to find the nearest counterfactual sample $\boldsymbol{x}_\text{cf}$ such that the outcome of classifier for $\boldsymbol{x}_\text{cf}$ changes to desired output class $y_\text{cf}$. In general, the counterfactual explanation $\boldsymbol{x}_\text{cf}$ for the individual $\boldsymbol{x}_\text{org}$ is the solution of the following optimization problem:


\begin{equation}
\label{eqn:original}
\boldsymbol{x}_\text{cf}^{*} = \argmin_{\boldsymbol{x}_\text{cf} \in \mathcal{X}} f(\boldsymbol{x}_\text{cf}) \quad\text{subject to}\quad \mathcal{H}(\boldsymbol{x}_\text{cf}^*) = y_\text{cf}
\end{equation}
\end{definition}
where $f(\boldsymbol{x}_\text{cf})$ is the function measuring the distance between $\boldsymbol{x}_\text{org}$ and $\boldsymbol{x}_\text{cf}$. Eq~\eqref{eqn:original} demonstrates the optimization objective that minimizes the similarity of the counterfactual and original samples, as well as ensures the classifier to change its decision output. For such explanations to be plausible, they should only suggest small changes in a few features.

To make it clear, we consider a simple scenario that a person with a set of features \{income: \$50k, CreditScore: ``good'', 
education: ``bachelor''
, age: 52\} applies for a loan in a financial organization and receives the reject decision from a predictive model. In this case, the company can utilize the counterfactual explanation (CF) as an advisor that provides constructive advice for this customer. To allow this customer successfully get the loan, CF can give an advice that how to change the customer's profile such as increasing his/her income to \$51k, or enhancing the education degree to ``Master''.
This toy example illustrates that CF is capable of providing interpretable advice that how to makes the least changes for the sample to achieve the desired outcome.

\subsection{Related Work}

Recently, there has been an increasing number of studies in this field. The existing counterfactual explanation methods can be categorized into gradient-based methods\cite{moore2019explaining,wachter2017counterfactual,mothilal2020explaining}, auto-encoder model \cite{dhurandhar2018explanations,mahajan2019preserving}, heuristic search based methods \cite{poyiadzi2020face,sharma2020certifai} and integer linear optimization \cite{cui2015optimal,kanamori2020dace}.


\textbf{Gradient-based methods:} Counterfactual explanation is first proposed by the study \cite{wachter2017counterfactual} as the example-based method to interpret machine learning models' decision. In this study, the authors construct the cross-entropy loss between the desired class and counterfactual samples' prediction with the purpose of changing the model output. Thereafter, some gradient-descent optimization algorithms would be used to minimize the constructed loss. This approach draws much attention with a plethora of studies \cite{grath2018interpretable,dhurandhar2018explanations,mothilal2020explaining,mothilal2020explaining} that aim to customize the loss function to enhance the properties of counterfactual generation. For example, the study~\cite{grath_interpretable_2018} extends the distance functions in Eq~\eqref{eqn:original} by using a weight vector ($\Theta$) to emphasize the importance of each feature. Some algorithms such as $k$-nearest neighbors or global feature evaluation can be deployed to find this vector ($\Theta$). Another framework called DiCE \cite{mothilal2020explaining} proposes using the diversity score to produce the number of generated samples that allows users to have more options. They thereafter use the weighted sum to combine different loss functions together and also adopt the gradient-descent algorithm to approximately find the optimal solution. The research \cite{van2019interpretable} utilizes the class prototype to guide the search progress to fall into the distribution of the expected class. This method however does not consider the causal relationship among features. The differentiable methods are the prominent approach in counterfactual explanation that allows to optimize easily and control the loss functions, but are only restricted to the differentiable models, and finds it hard to deal with the non-continuous values in tabular data. 


\textbf{Auto-encoder model:} Other recent studies based on the variational auto-encoder (VAE) model utilizes the properties of generative models to generate new counterfactual samples. In the study \cite{pawelczyk2020learning}, the authors first construct an encoder-decoder architecture. Thereafter, they generate the latent representation from the encoder, and make some perturbation into the latent representation, and go through the decoder until the prediction models achieve the desired class. Meanwhile, another line of recent work \cite{mahajan2019preserving} proposes the conditional auto-encoder model by combining different loss functions including prediction loss and proximity loss. They thereafter generate multiple counterfactual samples for all input data points by conditioning on the target class. These studies heavily rely on gradient-descent optimization which can face difficulties when handling categorical features. In addition, VAE models that maximize the lower bound of the log-likelihood rather than measuring the exact log-likelihood can give unstable and inconsistent results.



\textbf{Heuristic search methods:} There is an increasing number of counterfactual explanation methods for non-differentiable models, which makes the previous gradient-based approach not applicable. They utilizes heuristic search for the optimization problem such as Nelder-Mead \cite{grath2018interpretable}, growing spheres \cite{laugel2018comparison}, FISTA \cite{dhurandhar2019model,van2019interpretable}, or genetic algorithms \cite{dandl2020multi,lash2017generalized,sharma2020certifai}. The main idea of these approaches adopts evolutionary algorithms to effectively finds the optimal counterfactual samples based on the defined cost functions. For example, CERTIFAI \cite{sharma2020certifai} customizes the genetic algorithm for the counterfactuals search progress. CERTIFAI adopts the indicator functions (1 for different values, else 0) and mean squared error for categorical and continuous features, respectively. Apart from that, the study \cite{poyiadzi2020face} introduces a method called FACE that adopts Dijsstra's algorithm to generate counterfactual samples by finding the shortest path of the original input and the existing data points. The main advantage of FACE is that the produced path from Dijsstra's algorithm provides an insight into the step-by-step and feasible actions that users can take to achieve their goals. The generated samples of this method are limited to the input space without generating new data.



\textbf{Integer linear optimization} The studies \cite{ustun2019actionable,cui2015optimal} propose to adopt integer linear optimization (ILO) solver for linear models utilizing linear costs to generate the actionable changes. Specifically, they formulate the problem of finding counterfactual samples according to the cost function as a mixed-integer linear optimization problem and then utilize some existing solvers \cite{bliek1u2014solving} to obtain the optimal solution. To speed up the counterfactual samples search process, the study \cite{artelt2020convex} introduces convex constraints to bound the solutions in a region of data space locally. Although these approaches seem promising when dealing with non-continuous features and non-differentiable functions, they can be applied to linear models only.

Our method extends the line of studies \cite{van2019interpretable,mahajan2019preserving} by integrating both structural causal model and class prototype. We also formulate the problem as the multi-objective optimization problem and propose an algorithm to find the counterfactual samples effectively.

\section{Methodology}

\begin{figure}[!htb]
\centerline{\includegraphics[width=\textwidth]{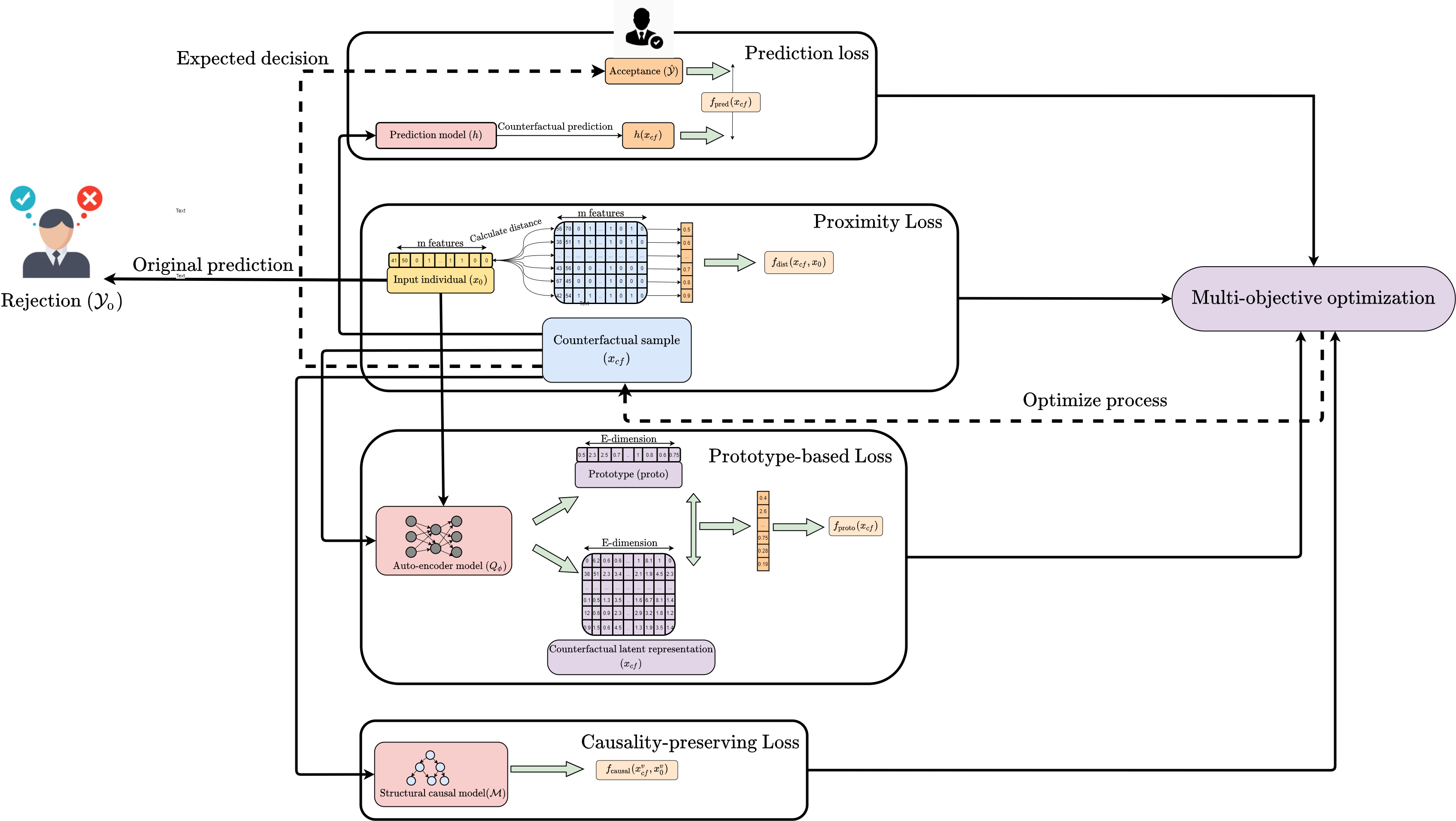}}
\caption{The overall framework for the proposed ProCE. The counterfactual samples are first initialized randomly.}
\label{fig:architecture}
\end{figure}

In this section, we firstly present different objective functions corresponding to different properties of counterfactual samples. The structural causal model and causal distance are also investigated to exploit the underlying causal relationship among features. Then, we formulate the counterfactual sample generation as a multi-objective optimization problem and propose an algorithm based on the non-dominated sorting genetic algorithm (NSGA-II) to obtain the optimal solutions. Figure~\ref{fig:architecture} generally describes the overall architecture of our proposed framework containing four main different loss functions: 1) prediction loss that ensures the valid counterfactual samples, 2) proximity loss encourages that only small changes would be performed in the counterfactual samples from the original one, 3) prototype-based loss that guides the search progress, and finally 4) causality-preserving loss that maintains the causal relationships. Moreover, there are three models in the framework: provided prediction model ($h$), auto-encoder model ($Q_\phi$), and structural causal model ($\mathcal{M}$).

\subsection{Prototype-based Causal Model}
\label{sec:objective}

Counterfactuals provide these explanations in the form of ``how to assign these features with different values, your credit application would have been accepted''. 
This indicates that counterfactual samples should be constrained under several particular conditions. We first provide definitions of each constraint condition and further tie them together as a multi-objective optimization problem to find an optimal counterfactual explanation. For clarity, we first introduce each constrain condition as loss function as follows.





\subsubsection{Prediction Loss}
We firstly consider the prediction loss which is the prominent loss function for counterfactual explanation. In order to achieve the desired outcome, prediction loss aims to calculate the distance between the counterfactual and expected/desired predictions. This loss function encourages the predictive models to change their predictions of counterfactual samples towards the desired outcomes. Particularly, for the classification scenario, we use the cross-entropy loss to minimize the counterfactual and expected outcome. The prediction loss is defined as follows:


\begin{equation}
\label{eqn:cross}
f_\text{pred}(\boldsymbol{x}_\text{cf}) = -{y_\text{cf}\log(\mathcal{H}(\boldsymbol{x}_\text{cf})) - (1 - y_\text{cf})\log(1 - \mathcal{H}(\boldsymbol{x}_\text{cf}))}
\end{equation}
Cross-entropy loss \cite{russell2019efficient} normally measures the performance of a classification model whose output is a probability value between 0 and 1. Cross-entropy loss is considered in this case to increases as the predicted probability of counterfactual samples $\mathcal{H}(\boldsymbol{x}_\text{cf})$ diverges from desired outcome $y_\text{cf}$. 

\subsubsection{Prototype-based Loss} 
\label{sec:proto}


In practice, the search space of counterfactuals might be incredibly large which thus results in slow optimization. Inspired by the work \cite{van2019interpretable}, we utilize the class prototype to guide the search progress with the aim of improving the efficiency of finding the counterfactual solutions. Class prototype is first defined as the mean encoding of the instances belonging to the same class \cite{snell2017prototypical}. Therefore, in our work, we construct an auto-encoder model to obtain the latent space which allows us to learn a better representation of these instances. 





We resort to an encoder function denoted by $Q_{\phi}: \mathcal{X} \rightarrow \mathbb{R}^E$
which projects the input feature $\mathcal{X}$ to the $E$-dimensional latent space. We denote $\mathcal{K}(\boldsymbol{x}_\text{org}) = \{ {\boldsymbol{x}_k, c_k}\}_{k=1}^K$ as a set of $K$-nearest instances of $\boldsymbol{x}_\text{org}$ by estimating the latent distance $||Q_{\phi}(\boldsymbol{x}_k) - Q_{\phi}(\boldsymbol{x}_\text{org})||^2_2$. 
Moreover, the classes 
of these $K$ instances, i.e., $\{c_k\}_{k=1}^K$, 
are different from the original prediction $y_\text{org}$ meaning that $c_k \neq y_\text{org}$.
Formally, $\mathcal{K}(\boldsymbol{x}_\text{org})$ is defined as:

\begin{equation}
    \mathcal{K}(\boldsymbol{x}_\text{org}) = \{ {\boldsymbol{x}_k, c_k}\}_{k=1}^K \subset \mathcal{D}
\end{equation}

such that

\begin{equation}
\label{eqn:distance}
\begin{cases}
c_k \neq y_\text{org}
\\
||Q_{\phi}(\boldsymbol{x}_r) - Q_{\phi}(\boldsymbol{x}_\text{org})||^2_2  \geq
||Q_{\phi}(\boldsymbol{x}_j) - Q_{\phi}(\boldsymbol{x}_\text{org})||^2_2 \quad \forall \boldsymbol{x}_r \in \{\mathcal{D} \backslash \mathcal{K}(\boldsymbol{x}_\text{org})\} 
\end{cases}
\end{equation}

Therefore, a prototype of an original instance $\boldsymbol{x}_\text{org}$ is computed by the mean of its nearest neighbors in the latent space:
\begin{equation}
    \text{proto}_*(\boldsymbol{x}_\text{org}) = \frac{1}{K}\sum_{\boldsymbol{x}_k \in \mathcal{K}(\boldsymbol{x}_\text{org})}{Q_{\phi}(\boldsymbol{x}_k)}
    \label{eq:proto_i}
\end{equation}

The definition of $\text{proto}_*$ in Eq.~\ref{eq:proto_i} 
indicates that the 
prototype is in fact the representatives of the samples belonging to counterfactual class.
We thus define the prototype loss function as $L_2$-norm distance between the representation of the counterfactual samples $\boldsymbol{x}_\text{cf}$ in the latent space and the obtained prototypes:
\begin{equation}
\label{eqn:protoloss}
    f_\text{proto}(\boldsymbol{x}_\text{cf}) = \| Q_{\phi}(\boldsymbol{x}_\text{cf}) - \text{proto}_* \|^2_2
\end{equation}

\subsubsection{Features cost}

One of the main obstacles of generating counterfactual samples is to compute the feature cost which captures the effort required for changing from original instance $\boldsymbol{x}_\text{org}$ to counterfactual ones $\boldsymbol{x}_\text{cf}$.
From the fundamental principles of counterfactual explanation, the generated samples should be as close as to the original one. The smallest changes mean that the least efforts are made for decision-makers to take to achieve their desired goals. However, even experts would find it hard to put the precise cost to demonstrate how unactionable the feature is. Moreover, when it comes to the mixed-type tabular data that contains both the categorical and continuous features, it is challenging to define the distance loss function \cite{jia2015new,kaufman2009finding,van2019distance,foss2019distance}. The previous studies \cite{sharma2020certifai,mothilal2020explaining,dandl2020multi} normally apply the indicator function that returns 1 when two categorical values match and returns 0 otherwise, and adopts $L_2$-norm distance for comparing continuous features. However, the indicator function which only returns 0 and 1 fails to measure the degree of similarity of two categories. In this study, we use the encoder model $Q_{\phi}$ to map the categorical features into the latent space before estimating their distance. The main advantage of this approach is that the encoder model has the capability to capture the underlying relationship and pattern between each categorical value. This means that manual feature engineering such as assigning weight for each category is not necessary, thus saving a great deal of time and effort. Thus, we come up with the distance between two samples is defined as below:

\begin{equation}
\label{eqn:distance}
f_\text{dist}(\boldsymbol{x}_\text{cf}, \boldsymbol{x}_{\text{org}}) = 
\begin{cases}
     \norm{\boldsymbol{x}^j_\text{cf} - \boldsymbol{x}^j_{\text{org}}}^2_2,& \text{if $\boldsymbol{x}^j$ is $j$-th continuous feature} \\
     \norm{Q_{\phi}(\boldsymbol{x}^j_\text{cf}) - Q_{\phi}(\boldsymbol{x}^j_\text{org})}^2_2,              &  \text{if $\boldsymbol{x}^j$ is $j$-th categorical feature}
\end{cases}
\end{equation}





\subsubsection{Causality-preserving Loss}

Although the distance function in Eq.~\eqref{eqn:distance} demonstrates the similarity of two samples, it fails to capture the causal relationship between each feature. To deal with this problem, we integrate the structural causal model, and thus construct the causal loss function to ensure the features' causal relationships in generated samples. We provide some fundamental definitions about causality and thereafter define the corresponding causal loss. In general, a structural causal model $\mathcal{M} = \{\mathbf{U}, \mathbf{V}, \mathbf{F}\}$ \cite{article_causal} consists of three main components defined as below:

\begin{itemize}
    \item $\mathbf{U}$ is the set of exogenous nodes which has no parents in the causal graph. 
    \item $\mathbf{V}$ is the set of random variables which are endogenous nodes whose causal mechanisms we are modeling. These variables have parents in the causal graph.
    \item $\mathbf{F}$ is the set of structural causal functions describing the causal relationships among the unobserved and observed variables. Specifically, for each node $\boldsymbol{X} \in V$, a function $f_X \in F$ such that $X = f_X(\text{Pa}(X), \mathbf{U}_X)$ where $\text{Pa}(X)$ is the parent nodes of $X$. 
\end{itemize}


A causal graph indicates a probabilistic graphical model that represents the assumptions about data-generating mechanism. A causal graph consists of a set of nodes and edges where each node represents a random variable, and each edge illustrates the causal relationship. The causal effect in causal model is facilitated by \textit{do-operator} or intervention \cite{pearl2000models} that assigns value $\boldsymbol{x}$ to a random variable $X$ denoted by $do(\boldsymbol{x})$. The symbol $do(x)$ is a model manipulation on a causal graph $\mathcal{M}$, which is defined as substitution of causal equation $X = f_X(\text{Pa}(X)_\mathcal{G}, \mathbf{U}_X)$ with $X=\boldsymbol{x}$. 


For each endogenous node $v \in V$, and its parent nodes (${v_{p_1}}, {v_{p_2}},\ldots, {v_{p_k}}$), we estimate each node $v$ as $v = g({v_{p_1}}, {v_{p_2}},\ldots, {v_{p_k}})$ to represent their causal relationship with $g(*)$ is the structural causal equation constructed by linear regression model. 
Since having the full causal graph is often impractical in real-world setting, it is quite challenging to estimate structural causal equation $g(*)$. In this work, we utilize LiNGAM \cite{shimizu2014lingam} which is a novel estimation technique based on the non-Gaussianity of the data to determine the function $g(*)$. During the counterfactuals generation progress, we firstly produce the predicted value of endogenous node $\boldsymbol{x}^v$ based on their parents before estimating the distance, which is measured as:

\begin{equation} 
\label{eqn:causal_dist}
\begin{split}
f_\text{causal}(\boldsymbol{x}_\text{cf}^{v}, \boldsymbol{x}_\text{org}^{v}) &= \norm{\boldsymbol{x}^{v}_\text{cf} - \boldsymbol{x}_\text{org}^{v}}^2_2 \\
&= \norm{g(\boldsymbol{x}^{v_{p_1}}_\text{cf}, \boldsymbol{x}^{v_{p_2}}_\text{cf},\ldots, \boldsymbol{x}^{v_{p_k}}_\text{cf}) - \boldsymbol{x}^{v}_{\text{org}}}^2_2
\end{split}
\end{equation}

With a set of observed variables containing the endogenous and exogenous ones $\boldsymbol{X} = \{\mathbf{U}, \mathbf{V}\}$, we can re-write the general distance between the original and counterfactual sample is the sum of distance of both of them. For the exogenous nodes $\mathbf{U}$ (nodes without any parents in the causal network), we still utilize the Eq.~\eqref{eqn:distance} which computing
the distance between two instances, while the causal distance in Eq.~\eqref{eqn:causal_dist} is employed for exogenous variables $\mathbf{V}$ (the remaining features).



\begin{equation}
\label{eqn:finaldist}
    f_{\text{final\_dist}}(\boldsymbol{x}_\text{cf}) = \sum_{u}^\mathbf{U} f_\text{dist}(\boldsymbol{x}^{u}_\text{cf}, \boldsymbol{x}^{u}_\text{org}) +  \sum_{v}^\mathbf{V} f_\text{causal}(\boldsymbol{x}^{v}_\text{cf}, \boldsymbol{x}^{v}_\text{org})
\end{equation}



\subsection{Multi-objective Optimization}

In this section, we aim to describe the proposed algorithm which is used for optimization process. With the loss functions presented in Sections~\ref{sec:objective} including $f_{\text{pred}}$, $f_{\text{proto}}$, $f_{\text{final\_dist}}$, we come up with the general objective functions Eq~\eqref{eq:obj}. These loss functions illustrates different properties that counterfactual samples should adhere to. The general loss functions containing three different losses is:

\begin{equation}
\label{eq:obj}
    \mathcal{L}(\boldsymbol{x}_\text{cf}) = \{f_\text{pred}(\boldsymbol{x}_\text{cf}),f_\text{proto}(\boldsymbol{x}_\text{cf}),f_\text{final\_dist}(\boldsymbol{x}_\text{cf})\}
\end{equation}

Therefore, the optimal solutions can be re-written as follows:

\begin{equation}
\label{eq:optimal}
    \boldsymbol{x}_\text{cf}^{*} = \argmin_{\boldsymbol{x}_\text{cf} \in \mathcal{X}} \mathcal{L}(\boldsymbol{x}_\text{cf})
\end{equation}

In order to obtain the optimal solutions, the majority of existing studies \cite{mahajan2019preserving,mothilal2020explaining,grath2018interpretable} uses the trade-off parameter sum assigning each loss function a weight, and combines them together. This approach seems to be reasonable; however, it is very challenging to balance the weights for each loss, resulting in a great deal of efforts and time into hyperparameter tuning. To address this issue, we propose to formulate the counterfactual explanation search as the multi-objective problem (MOP). In this study, we modify the elitist non-dominated sorting genetic algorithm (NSGA-II) \cite{deb2000fast} to deal with this optimization problem. Its main superiority is to optimize each loss function simultaneously as well as provide the solutions presenting the trade-offs among objective functions. To make it clear, we first present some related definitions. 
Given a set of $n$ candidate solutions $\mathcal{P} = \{\boldsymbol{x}_i\}_{i=1}^n$, we have the following ones:

\begin{definition}[Dominance in the objective space]
\label{def:dominance}
In the multi-objective optimization problem, the goodness of a solution is evaluated by the dominance\cite{deb2002fast}. Given two solutions $\boldsymbol{x}$ and $\boldsymbol{\hat{x}}$ along with a number of $p$ objective functions $f_i$, we have:
\begin{enumerate}
     \item $\boldsymbol{x}$ weakly dominates $\boldsymbol{\hat{x}}$ ($\boldsymbol{x} \succeq \boldsymbol{\hat{x}}$) iff $f_i(\boldsymbol{x}) \ge f_i(\boldsymbol{\hat{x}})$ $\forall i \in	\{1, \ldots , p\}$.
     \item $\boldsymbol{x}$ dominates $\boldsymbol{\hat{x}}$ ($\boldsymbol{x} \succ \boldsymbol{\hat{x}}$) iff $\boldsymbol{x} \succeq \boldsymbol{\hat{x}}$ and $\boldsymbol{x} \ne \boldsymbol{\hat{x}}$. 
\end{enumerate}     
\end{definition}

\begin{definition}[Pareto front]
\label{def:pareto}
Pareto front is a set of $m$ solutions denoted by ${F_*} = \{ \boldsymbol{x}_j \}_{j=1}^m \subset \mathcal{P}$  such that $\boldsymbol{x}_j$ dominates all remaining solutions $\boldsymbol{x}_r \in \{\mathcal{P} \backslash {{F}_*} \}$ with all objective functions. It means that $f_i(\boldsymbol{x}_j) \ge f_i(\boldsymbol{x}_r)$ $\forall i \in	\{1, \ldots , p\}$. The main goal of non-dominated solutions is to provide a reasonable compromise between all the objective functions that enhance one function's performance but not degrade others.



\end{definition}

\begin{definition}[Non-dominated sorting procedure]
\label{def:sorting}
Non-dominated sorting step is mainly used to sort the solutions in population according to the Pareto dominance principle, which plays a central role in the selection procedure. In fact, the set of candidate solutions $\mathcal{P}$ 
can be divided into a set of $H$ disjoint Pareto front as $\mathcal{F} = \{{F}_1,{F}_2,…,{F}_H\}$ where $H$ is the maximum number of fronts. Non-dominated sorting is a procedure for finding them. Particularly, in the non-dominated sorting step, all the non-dominated solutions from Definition~\ref{def:pareto} are selected from the population and are constructed as the Pareto front ${F}_1$. After that, the non-dominated solutions are chosen from the remaining population. The process is repeated until all the solutions are assigned to a front ${F}_H$.
\end{definition}


\begin{definition}[Crowding distance]
\label{def:crowding}
One of the vital characteristics of a population solution is diversity. In order to encourage the diversity of candidate solutions, the simplest approach is to choose the individuals having a low density. Particularly, to measure this characteristic, the crowding distance \cite{raquel2005effective,xu2017multi} is used to rank each candidate solution. Specficially, the crowding distance of an instance $\boldsymbol{x}$ is calculated as follows:


\begin{equation}
\label{eqn:crowding}
    d({\boldsymbol{x}}) = \sqrt{\sum_{i=1}^p \left(\frac{f_i(\boldsymbol{x}_a) - f_i(\boldsymbol{x}_b)}{f_i^\text{min} - f_i^\text{max}}\right)^2}
\end{equation}


where $p$ is the number of objective functions, $\boldsymbol{x}_a$ and $\boldsymbol{x}_b$ are two nearest instances of $\boldsymbol{x}$ by calculating the Euclidean distance,  $f_i$ is the $i$-th objective function, $f_i^\text{min}$ and $f_i^\text{max}$ are its minimum or maximum value, respectively. The fundamental concept behind crowding distance is to compute the Euclidean distance between each candidate solution $\{ \boldsymbol{x}_j \}_{j=1}^m$ in a front ${F_*}$
by using $p$ objective functions corresponding to $p$-dimensional hyper space.
\end{definition}

The optimization process for objective function~\eqref{eq:obj} is given by Algorithm~\ref{alg:mulobj}. The main idea behinds our approach is that for each generation, the algorithm chooses the Pareto Front for each objective function and evolves to the better ones. We firstly find the nearest class prototype of the original sample $\boldsymbol{x}_\text{org}$, which is used to measure the prototype loss function later. For the optimal counterfactual $\boldsymbol{x}_\text{cf}^*$ finding progress, each candidate solution is represented by the $D$-dimensional feature as the genes. A random candidate population is initialized with the Gaussian distribution. Thereafter, the objective functions including $f_{\text{pred}}$, $f_{\text{proto}}$, $f_{\text{final\_dist}}$ are calculated for each candidate. Non-dominated sorting procedure illustrated in Definition~\ref{def:sorting} is then performed to obtain a set of Pareto fronts ${\mathcal{F}} = \{{F}_i\}_{h=1}^H$.


The crowding distance function illustrated in Definition~\ref{def:crowding} and Eq.~\eqref{eqn:crowding} then is adopted as the score to assign to each individual in the current population. The algorithm only keeps the candidate solutions having the greatest ranking score, which illustrates that these solutions have low density. 
The cross-over and mutation procedures \cite{whitley1994genetic} are finally performed to generate the next population. Particularly, the cross-over of two parents generates the new candidate solutions by randomly swapping parts of genes. Meanwhile, the mutation procedure randomly alters some genes in the candidate solutions to encourage diversity and avoid local minimums. We repeat this process through many generations to find the optimal counterfactual solution.

    \scalebox{0.96}{
    \begin{minipage}{\linewidth}
\begin{algorithm}[H]
\small
\caption{Multi-objective Optimization for Prototype-based Counterfactual Explanation (ProCE)}
\label{alg:mulobj}
\begin{algorithmic}[1]
\renewcommand{\algorithmicrequire}{\textbf{Input:}}
\renewcommand{\algorithmicensure}{\textbf{Output:}}
\REQUIRE An original sample $\boldsymbol{x}_\text{org}$ with its prediction $y_\text{org}$, desired class $y_\text{cf}$, a provided machine learning classifier $\mathcal{H}$ and encoder model $Q_{\phi}$.
\STATE Compute prototype $\text{proto}_*$ by Eq.~\eqref{eq:proto_i}.
\STATE Initialize a batch of initial population with $n$ candidate solutions $\mathcal{P} =\{\boldsymbol{\Delta}_i\}_{i=1}^n$ with $\boldsymbol{\Delta}_i\sim\mathcal{N}(\boldsymbol{\mu},\boldsymbol{\nu})$.
\STATE $\mathcal{Q} = \emptyset$
\FOR{$g=1$ to $G$ generation}
\STATE $ \mathcal{P} = \mathcal{P} \cup \mathcal{Q}$
\FOR{each candidate solution $\boldsymbol{\Delta}_i$ in $\mathcal{P}$} 
\STATE Compute  $f_\text{pred}(\boldsymbol{\Delta}_i)$ based on Eq.~\eqref{eqn:cross}.
\STATE Use $\text{proto}_*$ to compute $f_\text{proto}(\boldsymbol{\Delta}_i)$ based on Eq.~\eqref{eqn:protoloss}.
\STATE Compute $f_\text{final\_dist}(\boldsymbol{\Delta}_i)$ based on Eq.~\eqref{eqn:finaldist}.
\ENDFOR
\STATE Obtain $\mathcal{F} = \{{F}_h\}_{h=1}^H$ by using non-dominated sorting procedure in Definition~\ref{def:sorting}. 
\STATE $\mathcal{P} = \emptyset$
\STATE $h = 0$ 
\WHILE{$|\mathcal{P}| + |{F}_h| < n$}
\STATE $\mathcal{P} = \mathcal{P} \cup {F}_h$
\STATE $h = h + 1$
\ENDWHILE

\STATE Compute the crowding distance as the ranking score for each solution in $\mathcal{P}$  based on Eq.~\eqref{eqn:crowding}.

\STATE Keep $n$ individuals in $\mathcal{P}$ based on ranking score. 

\STATE Randomly pair $\lceil n/2\rceil$  $\{\boldsymbol{\Delta}_1,\boldsymbol{\Delta}_2\}\in \mathcal{P}$
\FOR{each pair $\{\boldsymbol{\Delta}_1,\boldsymbol{\Delta}_2\}$}
\STATE Perform $\text{crossover}(\boldsymbol{\Delta}_1,\boldsymbol{\Delta}_2)\rightarrow \boldsymbol{\Delta}_1^{\prime},\boldsymbol{\Delta}_2^{\prime}$
\STATE Perform mutation  $\boldsymbol{\Delta}_1^{\prime}\rightarrow \tilde{\boldsymbol{\Delta}}_1,\boldsymbol{\Delta}_2^{\prime}\rightarrow \tilde{\boldsymbol{\Delta}}_2$
\STATE  $\mathcal{Q} = \mathcal{Q} \cup \{\tilde{\boldsymbol{\Delta}}_1,\tilde{\boldsymbol{\Delta}}_2\}$

\ENDFOR

\ENDFOR

\STATE $\boldsymbol{\Delta^*} \leftarrow \mathcal{P}[0]$
\ENSURE $\boldsymbol{x}_\text{cf} = \boldsymbol{\Delta^*}$
\end{algorithmic}
\end{algorithm}

    \end{minipage}%
    }

\section{Experiments}

We conduct experiments on four datasets to demonstrate the superior performance of our method when compared with state-of-the-art methods. All implementations are conducted in Python 3.7.7 with 64-bit Red Hat, Intel(R) Xeon(R) Gold 6150 CPU @ 2.70GHz. For our method, we construct the multi-objective optimization algorithm with the support of library Pymoo\footnote{\url{https://pymoo.org/algorithms/nsga2.html}} \cite{pymoo}. More details of implementation settings can be found in our code repository.




\subsection{Datasets}
\label{sec:dataset}
This section provides information about the datasets, on which we perform the comparison experiments. Our method is capable of generating counterfactual samples while maintaining the causal relationship. To validate this claim, we consider some feature conditions that restrict the generated counterfactual samples for each dataset. For simplicity, we denote $a \propto b$ for the condition that ($a$ \text{increase} $\Rightarrow$ $b$ \text{increase}) \text{AND} ($a$ \text{decrease} $\Rightarrow$ $b$ \text{decrease}). We use four datasets including \texttt{Simple-BN}, \texttt{Sangiovese}, \texttt{Adult} and \texttt{Law}.

    \texttt{Simple-BN} \cite{mahajan2019preserving} is a synthetic dataset containing 10,000 records with three features (${a}_1$,${a}_2$,${a}_3$) and a binary output ($y$). The data is generated based on the followed causal mechanism:
        \begin{equation}
        \label{eqn:scm}
        \begin{gathered}
        {a}_1 \sim \mathcal{N}(\mu_1, \sigma_1) \\ 
        {a}_2 \sim \mathcal{N}(\mu_2, \sigma_2) \\ 
        {a}_3 | {a}_1, {a}_2 \sim \mathcal{N}(k_3*({a}_1 + {a}_2)^2+b_3, \sigma_3) \\ 
        y | {a}_1, {a}_2, {a}_3 \sim \text{Ber}(\sigma(k_y*({a}_1*{a}_2)+b_y-{a}_3))
        \end{gathered}
        \end{equation}
        As illustrated by structural causal equations in Eq~\eqref{eqn:scm}, two random variables ${a}_1$ and ${a}_2$ follow the corresponding normal distribution $\mathcal{N}(\mu_1, \sigma_1)$ and $\mathcal{N}(\mu_2, \sigma_2)$, while ${a}_3$ follows the normal distribution with mean value determined by the function of ${a}_1$ and ${a}_2$. Additionally, target variable $y$ follows the Bernoulli distribution with the function of ${a}_1$, ${a}_2$ and ${a}_3$. Based on these generating mechanism, we consider the following causal relationship between ${a}_1$, ${a}_2$ and ${a}_3$:
    \begin{equation}
    ({a}_1, {a}_2) \propto {a}_3 
    \end{equation}

    The condition in Eq~\eqref{eqn:scm} means that ${a}_3$ monotonically increase and decrease by a function of two random variables ${a}_1$ and ${a}_2$. 
    


    \texttt{Sangiovese}\footnote{\url{https://www.bnlearn.com/bnrepository/clgaussian-small.html}}\cite{magrini2017conditional} dataset evaluates the impact of several agronomic settings on the quality of the Tuscan grapes. This dataset provides information about 14 continuous features along with the binary output. We consider the task of determining whether the grapes' quality is good or not. Based on the conditional linear Bayesian network provided with the dataset, we consider a causal relationship between two features including mean number of sprouts (SproutN) and mean number of bunches (BunchN) that is:


    \begin{equation}
        \text{BunchN} \propto \text{SproutN}
    \end{equation}

    \texttt{Adult}\footnote{\url{https://archive.ics.uci.edu/ml/datasets/adult}}\cite{Dua:2019} is the real-world dataset providing information of loan applicants in the financial organization. It is a mixed-type dataset that consists of instances having both continuous features and categorical features. For this dataset, we consider the task of determining whether the annual income of a person exceeds \$50k dollars.
    Similar to the study \cite{mahajan2019preserving}, with $\boldsymbol{x}_{*}^{\text{age}}$ and $\boldsymbol{x}_{*}^{\text{education}}$ referring to the feature age and education of an individual, we consider two conditions as below:
    \begin{equation}
         \boldsymbol{x}_\text{cf}^{\text{age}} \geq \boldsymbol{x}_\text{org}^{\text{age}}
    \end{equation}
    
    \begin{equation}
     \boldsymbol{x}_\text{cf}^\text{education} \propto \boldsymbol{x}_\text{org}^\text{age}
    \end{equation}
    
    Regarding the first condition ($\boldsymbol{x}_\text{cf}^{\text{age}} \geq \boldsymbol{x}_\text{org}^{\text{age}}$), counterfactual algorithms should not suggest decreasing individuals' ages since it violates the natural constraint that human age increases over time. Meanwhile, the second condition ($\boldsymbol{x}_\text{cf}^\text{education} \propto \boldsymbol{x}_\text{org}^\text{age}$) demonstrates the education-age causal relationship that obtaining a higher degree of education such as from ``Bachelor'' to ``PhD'' requires years to complete, thus causing age to increase. As a result, any counterfactual sample increasing education-level without increasing age is infeasible.

    \texttt{Law}\footnote{\url{http://www.seaphe.org/databases.php}}\cite{wightman1998lsac} dataset provides information of students with their features: sex, race and their entrance exam scores (LSAT), grade-point average (GPA) and first year average grade (FYA). The main task is to determine which applicants will be accepted to the law program. We consider a causal relationship:
    
    \begin{equation}
        (\text{LSAT}, \text{GPA}) \propto \text{FYA}
    \end{equation}
    
    In order to evaluate the models' effectiveness, we randomly split each dataset into 80\% training and 20\% test set. We conduct 100 repeated experiments, then evaluate performance on the test set and finally report the average statistics. 



\subsection{Evaluation Metrics}
\label{sec:eval}
In this section, we briefly describe six quantitative metrics that are used to evaluate the performance of our proposed method and baselines. We sample a number of $n$ factual samples and generate the counterfactual samples for them.  Meanwhile $n_{cat}$ and $n_{con}$ are the corresponding number of categorical and continuous features. $\mathbbm{1}(.)$ is the indicator function that returns 1 when the conditions are satisfied, otherwise returns 0. 


\textbf{Target-class validity} (\%Tcv) \cite{mahajan2019preserving,poyiadzi2020face} evaluates how well the algorithm can produce valid samples. Particularly, \%Tcv is calculated as the ratio of the number of samples belonging to the desired class and the number of factual samples. Higher target-class validity is favorable, demonstrating that the algorithm can generate greater numbers of counterfactual samples towards the desirable target variable.

\begin{equation}
    \text{\%Tcv} = \sum_{i=0}^{n} \frac{\mathbbm{1} (h(\boldsymbol{x}_\text{cf}) = y_\text{cf})}{n}
\end{equation}

\textbf{Causal-constraint validity} (\%Ccv) measures the percentage of counterfactual samples satisfying the pre-defined causal conditions. With this metric, the main aim is to evaluate how well our algorithm can generate feasible counterfactual samples that do not violate the causal relationship among features \cite{mahajan2019preserving}. With the causal conditions defined in the Section~\ref{sec:dataset}, using $n_s$ as the number samples satisfying causal conditions, the causal-constraint validity is defined in Eq~\eqref{eqn:ccv}. Higher causal-constraint validity is preferable, illustrating the greater number of satisfied counterfactual samples.

\begin{equation}
\label{eqn:ccv}
    \text{\%Ccv} =\frac{n_s}{n}
\end{equation}

\textbf{Categorical proximity} measures the proximity for categorical features representing the total number of matches on the values of each category between $\boldsymbol{x}_\text{cf}$ and $\boldsymbol{x}_\text{org}$. Higher categorical proximity is better, implying that the counterfactual sample preserves the minimal changes from the original  \cite{mothilal2020explaining}.

\begin{equation}
    \text{Cat\_proximity} = 1- \sum_{i=0}^{n} \sum_{j=0}^{{n_{cat}}} \mathbbm{1} (\boldsymbol{x}^j_\text{cf} \neq \boldsymbol{x}^j_\text{org})
\end{equation}


\textbf{Continuous proximity} illustrates the proximity of the continuous features, which is calculated as the negative of $L_2$-norm distance between the continuous features in $\boldsymbol{x}_
\text{cf}$ and $\boldsymbol{x}_\text{org}$. Higher continuous proximity is preferable, implying that the distance between the continuous features of $\boldsymbol{x}_\text{org}$ and $\boldsymbol{x}_\text{cf}$ should be as small as possible \cite{mothilal2020explaining}.

\begin{equation}
    \text{Con\_proximity} = -\sum_{i=0}^{n} \sum_{j=0}^{{n_{con}}} \norm{\boldsymbol{x}^j_{cf} - \boldsymbol{x}^j_0}^2_2
\end{equation}




\textbf{IM1 and IM2}
are two interpretability metrics (IM) proposed in \cite{van2019interpretable}. 
Let $Q^{\text{org}}_{\phi}$, $Q^{\text{cf}}_{\phi}$ and $Q^{\text{full}}_{\phi}$ be the auto-encoder models trained specifically on samples of class $y_\text{org}$, samples of class $y_\text{cf}$ and the full dataset, respectively, we first provide the general idea behind these two metrics. On the one hand, \textbf{IM1} measures the ratio of reconstruction errors of counterfactual sample $\boldsymbol{x}_\text{cf}$ using $Q^\text{cf}_{\phi}$ and  $Q^\text{org}_{\phi}$. A smaller value for \textbf{IM1} indicates that $\boldsymbol{x}_\text{cf}$ can be reconstructed more accurately by the autoencoder trained only on instances of the counterfactual class $y_\text{cf}$ than by the autoencoder trained on the original class $y_\text{org}$. This therefore demonstrate that the counterfactual sample $\boldsymbol{x}_\text{cf}$ lies closer to the data manifold of counterfactual class $y_\text{cf}$, which is considered to be more interpretable. On the other hand, \textbf{IM2} evaluates the similarity of counterfactual sample $\boldsymbol{x}_\text{cf}$ produced by $Q^\text{cf}_{\phi}$ and $Q_{\phi}$. A low value of IM2 means that the reconstructed instances of $\boldsymbol{x}_\text{cf}$ are very similar when using either $Q^\text{cf}_{\phi}$ or $Q^{\text{full}}_{\phi}$. Therefore, the data distribution of the counterfactual class $y_\text{cf}$ describes $x_\text{cf}$ as close as the distribution of all classes. Particularly, \textbf{IM1} and \textbf{IM2} are defined as follows:



\begin{equation}
    \text{IM1}(Q^\text{cf}_{\phi}, Q^\text{org}_{\phi}, \boldsymbol{x}_\text{cf}) = \sum_{i=0}^\text{n} \frac{\norm{\boldsymbol{x}_\text{cf} - Q^\text{cf}_{\phi}(\boldsymbol{x}_\text{cf})}^2_2}{\norm{ \boldsymbol{x}_\text{cf} - Q^\text{org}_{\phi}(\boldsymbol{x}_\text{cf})}^2_2  + \epsilon}
\end{equation}

\begin{equation}
    \text{IM2}(Q^\text{cf}_{\phi}, Q^\text{full}_{\phi}, \boldsymbol{x}_\text{cf}) = \sum_{i=0}^\text{n} \frac{\norm{ Q^\text{cf}_{\phi}(\boldsymbol{x}_\text{cf}) - Q^{\text{full}}_{\phi}(\boldsymbol{x}_\text{cf})}^^2_2}{\norm{\boldsymbol{x}_\text{cf}}^2_2 + \epsilon}
\end{equation}

\subsection{Baseline Methods} We compare our proposed method (ProCE) with several baselines including Wachter (AR), Growing Sphere (GS), CERTIFAI, CCHVAE and FACE. All of them are the recent approaches in the counterfactual explanation with available source codes and framework. The brief description of these baselines are illustrated as follows:




\begin{enumerate}

    \item \textbf{Wachter (Wach)} \cite{wachter2017counterfactual} which is a fundamental approach that generates counterfactual explanations by minimizing $L_1$-norm by using gradient descent to find counterfactuals $x_\text{cf}$ as close as to original instance $x_\text{org}$.

    \item \textbf{Growing Sphere (GS)} \cite{laugel2017inverse}
    is a random search algorithm, which generates samples around the factual input point until a point with a corresponding counterfactual class label was found. Growing hyperspheres are utilized to create the random samples around the original instance. This approach deals with immutable features by excluding them from the search procedure.
    \item \textbf{CERTIFAI} \cite{sharma2019certifai}
    CERTIFAI is an approach that utilizes genetic algorithm to finds the counterfactual samples more effectively. The source code for this method is not avaibale; therefore, we implement the CERTIFAI with the support from Python library PyGAD\footnote{\url{https://github.com/ahmedfgad/GeneticAlgorithmPython}}.
    \item \textbf{DiCE} \cite{mothilal2020explaining}. DiCE is one of the most prominent counterfactual explanation framework. This construct the weighted sum of different loss functions including proximity, diversity and sparsity together, and optimize the combined loss via the gradient-descent algorithm. For implementation, we utilize the source code\footnote{\url{https://github.com/divyat09/cf-feasibility}} with default settings.
    \item \textbf{FACE} \cite{poyiadzi2020face} produces a feasible and actionable set of counterfactual actions based on the shortest path lengths as determined by density-weighted metrics. The generated counterfactuals by this method that are plausible and coherent with the underlying data distribution.
    
\end{enumerate}


For all the experiments, we build two predictions model namely \textbf{1$^{\text{st}}$} \textbf{classifier} and \textbf{2$^{\text{nd}}$} \textbf{classifier}. The first classifier is a neural network with three hidden layers, while the second one has five hidden layers with the following architecture:

\textbf{1$^{\text{st}}$ classifier} \label{classifier1}
\begin{itemize}
    \item[$\bullet$] hidden Layer 1(Number of features, 64), batch normalization layer, dropout(0.1), activation function ReLU 
    \item[$\bullet$] hidden Layer 2(64, 32), batch normalization layer, Dropout(0.1), activation function ReLU 
    \item[$\bullet$] hidden Layer 3(32, 16), batch normalization layer, Dropout(0.1), activation function ReLU 
    \item[$\bullet$] last Layer (16, Data size), activation function sigmoid
\end{itemize}

\textbf{2$^{\text{nd}}$ classifier} \label{classifier2}
\begin{itemize}
    \item[$\bullet$] hidden layer 1(Number of features, 256), batch normalization layer, Dropout(0.1), activation function ReLU 
    \item[$\bullet$] hidden layer 2(Number of features, 128), batch normalization layer, Dropout(0.1), activation function ReLU 
    \item[$\bullet$] hidden layer 3(Number of features, 64), batch normalization layer, Dropout(0.1), activation function ReLU 
    \item[$\bullet$] hidden layer 4(64, 32), batch normalization layer, Dropout(0.1), activation function ReLU 
    \item[$\bullet$] hidden layer 5(32, 16), batch normalization layer, Dropout(0.1), activation function ReLU 
    \item[$\bullet$] last hidden layer (16, Data size), activation function sigmoid
\end{itemize}

The continuous features in datasets are in different value ranges; therefore, following the common practice in feature engineering~\cite{zheng2018feature,turner1999conceptual,dong2018feature}, we normalize the continuous feature to range (0,1). Moreover, regarding the categorical features, we transform them into numeric forms by using a label encoder.

\subsection{Results and Discussions}

 The performance of different metrics on {1$^\text{st}$} and {2$^\text{nd}$ classifier} are illustrated in Table~\ref{tab:result1} and ~\ref{tab:result2}, respectively. Regarding to the 1$^\text{st}$ classifier from Table~\ref{tab:result1}, all three methods achieve the competitive target-class validity, except the Watch performance in all datasets with around 90\% of samples belonging to the target class. Regarding the percentage of samples satisfying the causal constraints, by far the greatest performance is achieved by ProCE with 85.91\%, 91.84\%, 95.64\% and 90.43\% for \texttt{Simple-BN}, \texttt{Sangiovese}, \texttt{Adult} and \texttt{Law} datasets, respectively. FACE also produces a competitive performance across four datasets in terms of this metric, standing at 81.49\%, 88.65\%, 92.49\% and 86.71\% while the majority of generated samples from Watch violate the causal constraints (63.61\%, 58.1\%, 70.40\% and 76.71\%). The performance of \%Ccv cannot be achieved to 100\% for all the methods which demonstrates that it is quite challenging to maintain the causal constraints in counterfactual samples. Moreover, these results indicate that by integrating the structural causal model, our proposed method can effectively produce the counterfactual samples preserving the features' causal relationships. Regarding interpretability scores, our proposed method achieved the best IM1 and IM2 on four datasets. DiCE is ranked second recorded with competitive result in \texttt{Adult} dataset (0.0809 for IM1 and 0.2679 for IM2) and \texttt{Law} dataset (0.0423 for IM1 and 0.0427 for IM2). The performance of all metrics on the 2$^\text{nd}$ classifier in Table~\ref{tab:result2} also demonstrates the competitive performance of our proposed method across all metrics. We also notice that although the 2$^\text{nd}$ has a more complicated architecture than the 1$^\text{st}$ classifier, there is a small variation on the performance of counterfactual explanation algorithm. Finally, as expected, by using prototype as a guideline of the counterfactual search process, ProCE produces more interpretable counterfactual instances recorded with good performance in IM1 and IM2.
 By contrast, it is challenging for other approaches to reconstruct the counterfactual samples, leading to high interpretability scores (IM1 and IM2). 
 
On the other hand, to better comprehend the effectiveness of our proposed method in producing counterfactual samples compared with other approaches, we also perform a statistical significance test (paired $t$-test)
between our approach (ProCE) and other methods on each dataset and each metric with the obtained results on 100 randomly repeated experiments and report the result of $p$-value in Table~\ref{tab:result1} and ~\ref{tab:result2}. We find that our model is statistically significant with $p < 0.05$, thus demonstrating the effectiveness of ProCE in counterfactual samples generation task.

\begin{table*}[t]
\begin{adjustbox}{width=1.\columnwidth,center}
\begin{tabular}{@{}cccccccccc@{}}
\toprule
\multirow{2}{*}{\textbf{Method}} & \multirow{2}{*}{\textbf{Dataset}} & \multicolumn{4}{c}{\textbf{Performance}}                                                                                                            & \multicolumn{4}{c}{\textbf{$p$-value}}                                                                                                                \\ \cmidrule(l){3-6}  \cmidrule(l){7-10} 
                                 &                                   & \multicolumn{1}{c}{\textbf{\%Tcv}} & \multicolumn{1}{c}{\textbf{\%Ccv}} & \multicolumn{1}{c}{\textbf{IM1}} & \multicolumn{1}{c}{\textbf{IM2 (x10)}} & \textbf{\%Tcv}                      & \textbf{\%Ccv}                      & \textbf{IM1}                        & \textbf{IM2}                        \\ \midrule
Wach                             & \texttt{Simple-BN}                         & 91.00                              & 63.61                              & 0.0379  $\pm$    0.0741          & 0.0769  $\pm$    0.1385                & 0.0129                             & 0.0289                             & 0.0393                           & 0.0446                           \\
GS                               & \texttt{Simple-BN}                         & 100.00                             & 79.72                              & 0.0453    $\pm$  0.0835          & 0.0792    $\pm$  0.0202                & 0.0340                             & 0.0480                             & 0.0223                           & 0.0483                           \\
CERTIFAI                         & \texttt{Simple-BN}                         & 100.00                             & 77.44                              & 0.0489    $\pm$  0.1353          & 0.0271    $\pm$  0.0711                & 0.0098                             & 0.0226                             & 0.0365                           & 0.0218                           \\
DiCE                             & \texttt{Simple-BN}                         & 100.00                             & 73.61                              & 0.0376    $\pm$  0.1345          & 0.0815    $\pm$  0.1762                & 0.0227                             & 0.031                              & 0.0135                           & 0.0427                           \\
FACE                             & \texttt{Simple-BN}                         & 100.00                             & 81.49                              & 0.0365    $\pm$  0.0583          & 0.0429    $\pm$  0.1614                & 0.0256                             & 0.0197                             & 0.0444                           & 0.0468                           \\
\textbf{ProCE}                   & \texttt{Simple-BN}                         & \textbf{100.00}                    & \textbf{85.91}                     & \textbf{0.0322    $\pm$  0.1014} & \textbf{0.0211    $\pm$  0.0845}       & -              & -              & -            & -                                \\ \midrule
Wach                             & \texttt{Sangiovese}                        & 92.03                              & 58.10                              & 0.2513    $\pm$  0.1452          & 0.0533    $\pm$  0.0132                & 0.0260                             & 0.0365                             & 0.0447                           & 0.0358                           \\
GS                               & \texttt{Sangiovese}                        & 100.00                             & 89.60                              & 0.2295    $\pm$  0.0584          & 0.0425    $\pm$  0.1502                & 0.0131                             & 0.0469                             & 0.014                            & 0.0162                           \\
CERTIFAI                         & \texttt{Sangiovese}                        & 100.00                             & 74.29                              & 0.2915    $\pm$  0.1920          & 0.0721    $\pm$  0.1366                & 0.0410                             & 0.0389                             & 0.0215                           & 0.0212                           \\
DiCE                             & \texttt{Sangiovese}                        & 100.00                             & 78.10                              & 0.2447    $\pm$  0.0759          & 0.0374    $\pm$  0.1657                & 0.0297                             & 0.0306                             & 0.0388                           & 0.0102                           \\
FACE                             & \texttt{Sangiovese}                        & 100.00                             & 88.65                              & 0.2424    $\pm$  0.0962          & 0.0873    $\pm$  0.0495                & 0.0471                             & 0.0148                             & 0.0140                           & 0.0119                           \\
\textbf{ProCE}                   & \texttt{Sangiovese}                        & \textbf{100.00}                    & \textbf{91.84}                     & \textbf{0.2152    $\pm$  0.1686} & \textbf{0.0370    $\pm$  0.0574}       & -              & -              & -            & -                                \\ \midrule
Wach                             & \texttt{Adult}                             & 93.95                              & 70.40                              & 0.0709    $\pm$  0.1582          & 0.3063    $\pm$  0.1382                & 0.048                              & 0.0285                             & 0.0242                           & 0.0407                           \\
GS                               & \texttt{Adult}                             & 100.00                             & 70.13                              & 0.2241    $\pm$  0.0396          & 0.3343    $\pm$  0.0564                & 0.0144                             & 0.0274                             & 0.0114                           & 0.0468                           \\
CERTIFAI                         & \texttt{Adult}                             & 100.00                             & 91.99                              & 0.0939    $\pm$  0.0834          & 0.3735    $\pm$  0.1150                & 0.0320                             & 0.0348                             & 0.0310                           & 0.0222                           \\
DiCE                             & \texttt{Adult}                             & 100.00                             & 80.40                             & 0.0809    $\pm$  0.1538          & 0.2679    $\pm$  0.1661                & 0.0318                             & 0.0169                             & 0.0275                           & 0.0415                           \\
FACE                             & \texttt{Adult}                             & 100.00                             &     92.49                           & 0.1283    $\pm$  0.0336          & 0.3245    $\pm$  0.1881                & 0.0215                             & 0.0346                             & 0.019                            & 0.0242                           \\
\textbf{ProCE}                   & \texttt{Adult}                             & \textbf{100.00}                    & \textbf{95.64}                     & \textbf{0.0675    $\pm$  0.1908} & \textbf{0.2171    $\pm$  0.0546}       & -              & -              & -            & -                                \\ \midrule
Wach                             & \texttt{Law}                               & 92.45                              & 76.71                              & 0.0536    $\pm$  0.1312          & 0.0470    $\pm$  0.0800                & 0.0159                             & 0.026                              & 0.0115                           & 0.0378                           \\
GS                               & \texttt{Law}                               & 100.00                             & 86.23                              & 0.0484    $\pm$  0.1173          & 0.0487    $\pm$  0.0858                & 0.0481                             & 0.0392                             & 0.0314                           & 0.0315                           \\
CERTIFAI                         & \texttt{Law}                               & 100.00                             & 82.72                              & 0.0567    $\pm$  0.1427          & 0.0461    $\pm$  0.1797                & 0.0102                             & 0.0425                             & 0.0191                           & 0.0340                           \\
DiCE                             & \texttt{Law}                               & 100.00                             & 85.75                              & 0.0423    $\pm$  0.1902          & 0.0427    $\pm$  0.0801                & 0.0138                             & 0.0206                             & 0.0122                           & 0.0487                           \\
FACE                             & \texttt{Law}                               & 100.00                             & 86.71                              & 0.0418     $\pm$  0.0125          & 0.0435    $\pm$  0.1160                & 0.0125                             & 0.0315                             & 0.0333                           & 0.0450                           \\
\textbf{ProCE}                   & \texttt{Law}                               & \textbf{100.00}                    & \textbf{90.43}                     & \textbf{0.0410    $\pm$  0.1268} & \textbf{0.0421    $\pm$  0.1907}       & -              & -              & -            & -            \\ \bottomrule

\end{tabular}
\end{adjustbox}
\caption{Performance of all methods on $1^\text{st}$ classifier.
We compute $p$-value by conducting a paired $t$-test between our approach (ProCE) and baselines with 100 repeated experiments for each metric. }
\label{tab:result1}
\end{table*}

\begin{table*}[t]
\begin{adjustbox}{width=1.\columnwidth,center}
\begin{tabular}{@{}cccccccccc@{}}
\toprule
\multirow{2}{*}{\textbf{Method}} & \multirow{2}{*}{\textbf{Dataset}} & \multicolumn{4}{c}{\textbf{Performance}}                                                                                                            & \multicolumn{4}{c}{\textbf{$p$-value}}                                                                                                                \\ \cmidrule(l){3-6}  \cmidrule(l){7-10} 
                                 &                                   & \multicolumn{1}{c}{\textbf{\%Tcv}} & \multicolumn{1}{c}{\textbf{\%Ccv}} & \multicolumn{1}{c}{\textbf{IM1}} & \multicolumn{1}{c}{\textbf{IM2 (x10)}} & \textbf{\%Tcv}                      & \textbf{\%Ccv}                      & \textbf{IM1}                        & \textbf{IM2}                        \\ \midrule
Wach                             & \texttt{Simple-BN}                         & 93.33                              & 70.96                              & 0.0512  $\pm$    0.0466          & 0.0262  $\pm$    0.0507                & 0.0320                             & 0.0096                             & 0.0372                           & 0.0487                           \\
GS                               & \texttt{Simple-BN}                         & 100.00                             & 79.46                              & 0.0401    $\pm$  0.1888          & 0.0354    $\pm$  0.0352                & 0.0242                             & 0.038                              & 0.0274                           & 0.0308                           \\
CERTIFAI                         & \texttt{Simple-BN}                         & 100.00                             & 83.68                              & 0.0465    $\pm$  0.0389          & 0.0824    $\pm$  0.1345                & 0.0378                             & 0.0138                             & 0.031                            & 0.0255                           \\
DiCE                             & \texttt{Simple-BN}                         & 100.00                             & 82.93                              & 0.0342    $\pm$  0.0790          & 0.0448    $\pm$  0.0260                & 0.0376                             & 0.0324                             & 0.0497                           & 0.0277                           \\
FACE                             & \texttt{Simple-BN}                         & 100.00                             & 82.03                              & 0.0458    $\pm$  0.1209          & 0.0435    $\pm$  0.0123                & 0.0215                             & 0.0086                             & 0.0275                           & 0.0437                           \\
ProCE                            & \texttt{Simple-BN}                         & \textbf{100.00}                    & \textbf{89.09}                     & \textbf{0.0318    $\pm$  0.0104} & \textbf{0.0202    $\pm$  0.0167}       & -              & -              & -            & -            \\ \midrule
Wach                             & \texttt{Sangiovese}                        & 93.92                              & 74.49                              & 0.2731    $\pm$  0.1090          & 0.0445    $\pm$  0.0919                & 0.0255                             & 0.0291                             & 0.0474                           & 0.0363                           \\
GS                               & \texttt{Sangiovese}                        & 100.00                             & 71.44                              & 0.2654    $\pm$  0.0394          & 0.0407    $\pm$  0.0770                & 0.0319                             & 0.0378                             & 0.0294                           & 0.0447                           \\
CERTIFAI                         & \texttt{Sangiovese}                        & 100.00                             & 80.95                              & 0.2583    $\pm$  0.1369          & 0.0798    $\pm$  0.1898                & 0.0281                             & 0.0304                             & 0.0389                           & 0.0297                           \\
DiCE                             & \texttt{Sangiovese}                        & 100.00                             & 92.25                              & 0.2603    $\pm$  0.1383          & 0.0880    $\pm$  0.1144                & 0.0436                             & 0.0323                             & 0.0478                           & 0.0381                           \\
FACE                             & \texttt{Sangiovese}                        & 100.00                             & 77.95                              & 0.2302    $\pm$  0.0029          & 0.0522    $\pm$  0.0169                & 0.0464                             & 0.0152                             & 0.0351                           & 0.0184                           \\
ProCE                            & \texttt{Sangiovese}                        & \textbf{100.00}                    & \textbf{86.25}                     & \textbf{0.2127    $\pm$  0.0973} & \textbf{0.0360    $\pm$  0.0388}       & -              & -              & -            & -            \\ \midrule
Wach                             & \texttt{Adult}                             & 91.45                              & 75.23                              & 0.1731    $\pm$  0.1270          & 0.3520    $\pm$  0.1592                & 0.0127                             & 0.0454                             & 0.0407                           & 0.0378                           \\
GS                               & \texttt{Adult}                             & 100.00                             & 75.82                              & 0.1719    $\pm$  0.1673          & 0.1565    $\pm$  0.1634                & 0.0308                             & 0.0099                             & 0.0224                           & 0.0447                           \\
CERTIFAI                         & \texttt{Adult}                             & 100.00                             & 80.56                              & 0.1512    $\pm$  0.0920          & 0.2326    $\pm$  0.0686                & 0.0265                             & 0.0351                             & 0.0309                           & 0.0341                           \\
DiCE                             & \texttt{Adult}                             & 100.00                             & 76.43                              & 0.2371    $\pm$  0.1801          & 0.3823    $\pm$  0.0016                & 0.0154                             & 0.0396                             & 0.0427                           & 0.0343                           \\
FACE                             & \texttt{Adult}                             & 100.00                             & 76.02                              & 0.1649    $\pm$  0.1448          & 0.3393    $\pm$  0.0083                & 0.0254                             & 0.0144                             & 0.0105                           & 0.0285                           \\
ProCE                            & \texttt{Adult}                             & \textbf{100.00}                    & \textbf{92.85}                     & \textbf{0.1467    $\pm$  0.1096} & \textbf{0.1324    $\pm$  0.1027}       & -              & -              & -            & -            \\ \midrule
Wach                             & \texttt{Law}                               & 90.55                              & 73.36                              & 0.0437    $\pm$  0.0913          & 0.0594    $\pm$  0.1896                & 0.0375                             & 0.0474                             & 0.0462                           & 0.0349                           \\
GS                               & \texttt{Law}                               & 100.00                             & 84.09                              & 0.0532    $\pm$  0.0988          & 0.0643    $\pm$  0.0244                & 0.0269                             & 0.0267                             & 0.0402                           & 0.0334                           \\
CERTIFAI                         & \texttt{Law}                               & 100.00                             & 80.88                              & 0.0382    $\pm$  0.0915          & 0.0592    $\pm$  0.0566                & 0.0495                             & 0.0172                             & 0.0428                           & 0.0286                           \\
DiCE                             & \texttt{Law}                               & 100.00                             & 87.54                              & 0.0382    $\pm$  0.0530          & 0.0461    $\pm$  0.1928                & 0.0421                             & 0.0489                             & 0.0342                           & 0.0373                           \\
FACE                             & \texttt{Law}                               & 100.00                             & 75.51                              & 0.0422    $\pm$  0.1875          & 0.0383    $\pm$  0.0029                & 0.0476                             & 0.0374                             & 0.015                            & 0.0304                           \\
ProCE                            & \texttt{Law}                               & \textbf{100.00}                    & \textbf{79.48}                     & \textbf{0.0317    $\pm$  0.1073} & \textbf{0.0313    $\pm$  0.1648}       & -              & -              & -            & -            \\ \bottomrule      

\end{tabular}
\end{adjustbox}
\caption{Performance of all methods on $2^\text{nd}$ classifier. 
We compute $p$-value by conducting a paired $t$-test between our approach (ProCE) and baselines with 100 repeated experiments for each metric. }
\label{tab:result2}
\end{table*}

\begin{table*}[!htb]
\begin{adjustbox}{width=0.6\columnwidth,center}
\begin{tabular}{@{}cccc@{}}

\toprule
\multirow{2}{*}{\textbf{Method}} & \multirow{2}{*}{\textbf{Dataset}} & \multicolumn{2}{c}{\textbf{Running time (s)}}                    \\ \cmidrule(l){3-4} 
                                 &                                   & \textbf{1$^\text{st}$ classifier}  & \textbf{2$^\text{nd}$ classifier} \\ \cmidrule(r){1-4}
Wach     & \texttt{Simple-BN}  & \textbf{3.030  $\pm$    0.105} & \textbf{5.111  $\pm$    0.135} \\
GS       & \texttt{Simple-BN}  & 7.126    $\pm$  0.153          & 6.541    $\pm$  0.053          \\
CERTIFAI & \texttt{Simple-BN}  & 6.213    $\pm$  0.007          & 6.237    $\pm$  0.088          \\
DiCE     & \texttt{Simple-BN}  & 6.522    $\pm$  0.088          & 6.455    $\pm$  0.016          \\
FACE     & \texttt{Simple-BN}  & 8.022    $\pm$  0.014          & 6.599    $\pm$  0.173          \\
ProCE    & \texttt{Simple-BN}  & 4.085    $\pm$  0.055          & 6.017    $\pm$  0.160          \\ \midrule
Wach     & \texttt{Sangiovese} & \textbf{5.125    $\pm$  0.097} & \textbf{5.768    $\pm$  0.113} \\
GS       & \texttt{Sangiovese} & 8.048    $\pm$  0.176          & 12.549    $\pm$  0.086         \\
CERTIFAI & \texttt{Sangiovese} & 7.688    $\pm$  0.131          & 8.906    $\pm$  0.105          \\
DiCE     & \texttt{Sangiovese} & 13.426    $\pm$  0.158         & 11.775    $\pm$  0.086         \\
FACE     & \texttt{Sangiovese} & 7.810    $\pm$  0.076          & 11.348    $\pm$  0.200         \\
ProCE    & \texttt{Sangiovese} & 6.809    $\pm$  0.162          & 7.304    $\pm$  0.101          \\ \midrule
Wach     & \texttt{Adult}      & 7.046    $\pm$  0.151          & 7.260    $\pm$  0.058          \\
GS       & \texttt{Adult}      & 6.472    $\pm$  0.021          & 6.464    $\pm$  0.145          \\
CERTIFAI & \texttt{Adult}      & 13.851    $\pm$  0.001         & 9.457    $\pm$  0.120          \\
DiCE     & \texttt{Adult}      & 7.943    $\pm$  0.046          & 10.326    $\pm$  0.016         \\
FACE     & \texttt{Adult}      & 10.821    $\pm$  0.162         & 9.140    $\pm$  0.149          \\
ProCE    & \texttt{Adult}      & \textbf{4.837    $\pm$  0.026} & \textbf{5.733    $\pm$  0.019} \\ \midrule
Wach     & \texttt{Law}        & \textbf{4.821    $\pm$  0.068} & \textbf{4.957    $\pm$  0.131} \\
GS       & \texttt{Law}        & 12.126    $\pm$  0.093         & 13.480    $\pm$  0.152         \\
CERTIFAI & \texttt{Law}        & 5.516    $\pm$  0.009          & 6.337    $\pm$  0.027          \\
DiCE     & \texttt{Law}        & 6.150    $\pm$  0.038          & 8.103    $\pm$  0.0410         \\
FACE     & \texttt{Law}        & 5.450    $\pm$  0.184          & 6.661    $\pm$  0.025          \\
ProCE    & \texttt{Law}        & 4.830    $\pm$  0.130          & 5.001   $\pm$  0.152 \\ \bottomrule
\end{tabular}
\end{adjustbox}
\caption{We report running time of different methods on four datasets.}
\label{tab:runningtime}
\end{table*}

\begin{figure}[!htb]
\centerline{\includegraphics[width=0.7\textwidth]{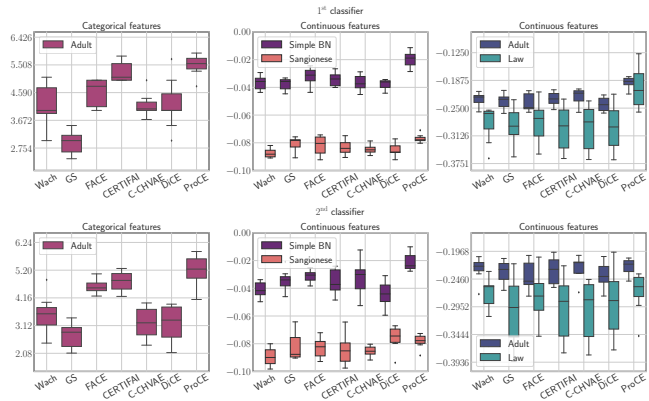}}
\caption{Baseline results in terms of \textbf{Continuous proximity} and \textbf{Categorical proximity}. Higher continuous and categorical proximity are better.}
\label{fig:proximity}
\end{figure}

Figure~\ref{fig:proximity} provides information about the categorical proximity in the Adult dataset and continuous proximity in four datasets. For the categorical proximity on both 1$^\text{st}$ and 2$^\text{nd}$ classifier, ProCE consistently achieves an average of 5 out of the total 6 categories in the dataset meaning that the counterfactual sample generated from ProCE preserves an average of 5 categorical features from the original instances. CERTIFAI and FACE also yield competitive results for categorical proximity, whereas the lowest result is recorded in the GS algorithm (1.7 to 3.5 categories). These results illustrate that the gradient-free based approach including ProCE, CERTIFAI and FACE can achieve better performance in handling the non-continuous features in tabular data. When it comes to the continuous proximity, ProCE produces the counterfactual sample with the greatest similarity around over -0.02, -0.078, -0.1875 and -0.17 corresponding to \texttt{Simple-BN}, \texttt{Sangiovese}, \texttt{Adult} and \texttt{Law dataset}. Our proposed method produces the least fluctuation in continuous proximity for \texttt{Sangiovese}, \texttt{Simple-BN}, \texttt{Adult}, while the biggest variation is witnessed in \texttt{Law}. 


We also report the running time of different methods in Table~\ref{tab:runningtime}. Overall, the shortest time is recorded with Watch method on \texttt{Simple-BN}, \texttt{Sangiovese}, and \texttt{Law} datasets. The possible reason is that Watch is the naive approach which optimizes the basic proximity loss functions using gradient descent. This therefore allows producing the counterfactual sample in a prominent time but demonstrates a poor performance in several metrics. Our approach (ProCE) also demonstrates competitive time performance on these three datasets. Regarding \texttt{Adult} dataset which contains both categorical and continuous features, our approach performs counterfactual sample generation in the outstanding time and also surpasses other non-gradient descent methods such as FACE, CERTIFAI and GS.

\begin{figure}
\centering
\begin{subfigure}[b]{\textwidth}
  \includegraphics[width=1\linewidth]{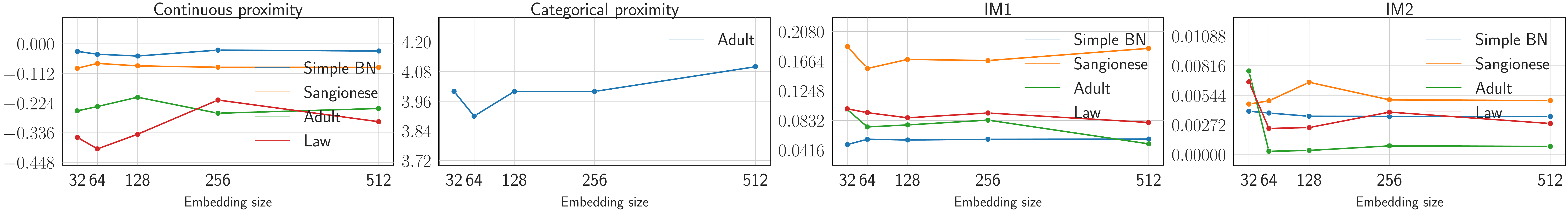}
  \caption{Our performance under different sizes of $E$-dimensional embedding for encoder function $Q_{\phi}$.}
  \label{fig:size} 
\end{subfigure}
\begin{subfigure}[b]{\textwidth}
  \includegraphics[width=1\linewidth]{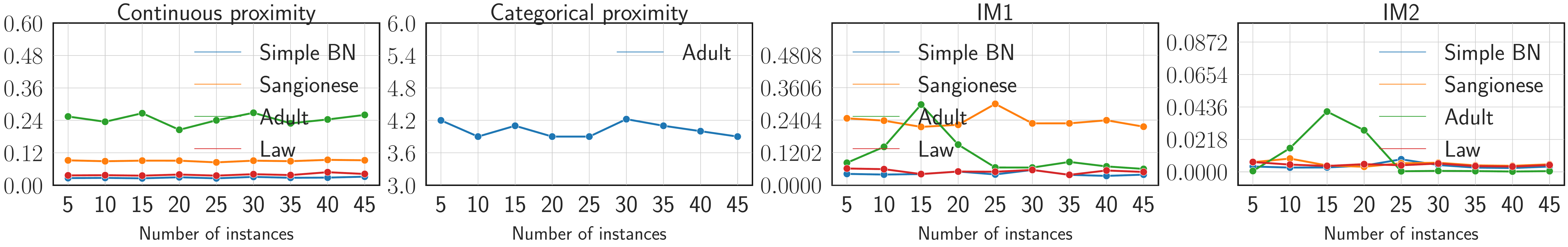}
  \caption{Our performance under different numbers of $K$-nearest neighbors for class prototype }
  \label{fig:instance}
\end{subfigure}
\caption[Two numerical solutions]{Sensitivity of hyperparameters. }
\end{figure}

Figure~\ref{fig:size} and~\ref{fig:instance} show the variation of our method's performance with the different numbers of $K$-nearest neighbors for class prototype and $E$-dimensional embedding sizes of the auto-encoder model, respectively. It is clear from Figure~\ref{fig:size} that the performance of continuous proximity for \texttt{Simple-BN}, \texttt{Sangiovese} and \texttt{Adult} datasets is nearly stable with different embedding sizes, while \texttt{Law} witnesses a quite significant variation, increasing from around -0.336 to -0.224 corresponding to embedding sizes of 32 to 256, followed by a slight decrease to -0.33 (embedding size 512). A similar pattern also is recorded for the remaining metrics including categorical proximity, IM1, and IM2 with the good and stable performance at an embedding size of 256. The slight small fluctuations possibly illustrate that the impact of embedding size on the model performance is not very significant. Moreover, 256 is the preferable embedding size, while the sizes of 32 and 512 seem to be relatively small and large to sufficiently capture latent information for embedding vectors. Regarding categorical proximity, the performance declines slightly by 0.1 from 32 to 64, and thereafter varies slightly around 4.0 - 4.09 with embedding sizes of 128, 256, and 512. On the other hand, as can be seen from Figure~\ref{fig:instance}, IM1 and IM2 demonstrate a similar pattern illustrated by the worst performance when the number of instances of 15, followed by a stagnant performance from 25 to 45 instances. It is believed that the similar trend occurring in IM1 and IM2 is reasonable due to their similar properties illustrated in Section~\ref{sec:eval}. Meanwhile, there is no significant variation in the performance of continuous and categorical proximity across four datasets. These results suggest that the performance of our proposed method witnesses a small variation in all evaluation metrics regarding two hyperparameters (embedding sizes and numbers of nearest neighbors), implying our model's stability and robustness.

\section{Conclusion}
This paper introduces a novel counterfactual explanation algorithm by integrating the structural causal model and the class prototype. We also proposed formulating the counterfactual generation as a multi-objective problem and construct an optimization algorithm to find the optimal counterfactual explanation in an effective manner. Our experiments validate that our method outperforms the state-of-the-art methods on many evaluation metrics. For future work, we plan to extend our framework to the imperfect structural causal model that is very commonplace in real-world scenarios. Meanwhile, other multi-objective optimization algorithms such as reinforcement learning and multi-task learning are also worthy of investigation.

\bibliography{www} 

\begin{thebibliography}{10}
\expandafter\ifx\csname url\endcsname\relax
  \def\url#1{\texttt{#1}}\fi
\expandafter\ifx\csname urlprefix\endcsname\relax\def\urlprefix{URL }\fi
\expandafter\ifx\csname href\endcsname\relax
  \def\href#1#2{#2} \def\path#1{#1}\fi

\bibitem{zavrvsnik2019algorithmic}
A.~Zavr{\v{s}}nik, Algorithmic justice: Algorithms and big data in criminal
  justice settings, European Journal of Criminology (2019) 1477370819876762.

\bibitem{kaur2020interpreting}
H.~Kaur, H.~Nori, S.~Jenkins, R.~Caruana, H.~Wallach, J.~Wortman~Vaughan,
  Interpreting interpretability: Understanding data scientists' use of
  interpretability tools for machine learning, in: Proceedings of the 2020 CHI
  Conference on Human Factors in Computing Systems, 2020, pp. 1--14.

\bibitem{galindo2000credit}
J.~Galindo, P.~Tamayo, Credit risk assessment using statistical and machine
  learning: basic methodology and risk modeling applications, Computational
  Economics 15~(1) (2000) 107--143.

\bibitem{schwab2019cxplain}
P.~Schwab, W.~Karlen, Cxplain: Causal explanations for model interpretation
  under uncertainty, arXiv preprint arXiv:1910.12336.

\bibitem{williams2016axis}
J.~J. Williams, J.~Kim, A.~Rafferty, S.~Maldonado, K.~Z. Gajos, W.~S. Lasecki,
  N.~Heffernan, Axis: Generating explanations at scale with learnersourcing and
  machine learning, in: Proceedings of the Third (2016) ACM Conference on
  Learning@ Scale, 2016, pp. 379--388.

\bibitem{zhao2021causal}
Q.~Zhao, T.~Hastie, Causal interpretations of black-box models, Journal of
  Business \& Economic Statistics 39~(1) (2021) 272--281.

\bibitem{ustun2019actionable}
B.~Ustun, A.~Spangher, Y.~Liu, Actionable recourse in linear classification,
  in: Proceedings of the Conference on Fairness, Accountability, and
  Transparency, 2019, pp. 10--19.

\bibitem{poyiadzi2020face}
R.~Poyiadzi, K.~Sokol, R.~Santos-Rodriguez, T.~De~Bie, P.~Flach, Face: feasible
  and actionable counterfactual explanations, in: Proceedings of the AAAI/ACM
  Conference on AI, Ethics, and Society, 2020, pp. 344--350.

\bibitem{sharma2020certifai}
S.~Sharma, J.~Henderson, J.~Ghosh, Certifai: A common framework to provide
  explanations and analyse the fairness and robustness of black-box models, in:
  Proceedings of the AAAI/ACM Conference on AI, Ethics, and Society, 2020, pp.
  166--172.

\bibitem{dhurandhar2019model}
A.~Dhurandhar, T.~Pedapati, A.~Balakrishnan, P.-Y. Chen, K.~Shanmugam, R.~Puri,
  Model agnostic contrastive explanations for structured data, arXiv preprint
  arXiv:1906.00117.

\bibitem{grath2018interpretable}
R.~M. Grath, L.~Costabello, C.~L. Van, P.~Sweeney, F.~Kamiab, Z.~Shen,
  F.~Lecue, Interpretable credit application predictions with counterfactual
  explanations, arXiv preprint arXiv:1811.05245.

\bibitem{lash2017generalized}
M.~T. Lash, Q.~Lin, N.~Street, J.~G. Robinson, J.~Ohlmann, Generalized inverse
  classification, in: Proceedings of the 2017 SIAM International Conference on
  Data Mining, SIAM, 2017, pp. 162--170.

\bibitem{mahajan2019preserving}
D.~Mahajan, C.~Tan, A.~Sharma, Preserving causal constraints in counterfactual
  explanations for machine learning classifiers, arXiv preprint
  arXiv:1912.03277.

\bibitem{mothilal2020explaining}
R.~K. Mothilal, A.~Sharma, C.~Tan, Explaining machine learning classifiers
  through diverse counterfactual explanations, in: Proceedings of the 2020
  Conference on Fairness, Accountability, and Transparency, 2020, pp. 607--617.

\bibitem{fernandez2020explaining}
C.~Fern{\'a}ndez-Lor{\'\i}a, F.~Provost, X.~Han, Explaining data-driven
  decisions made by ai systems: The counterfactual approach, arXiv preprint
  arXiv:2001.07417.

\bibitem{moore2019explaining}
J.~Moore, N.~Hammerla, C.~Watkins, Explaining deep learning models with
  constrained adversarial examples, in: Pacific Rim International Conference on
  Artificial Intelligence, Springer, 2019, pp. 43--56.

\bibitem{wachter2017counterfactual}
S.~Wachter, B.~Mittelstadt, C.~Russell, Counterfactual explanations without
  opening the black box: Automated decisions and the gdpr, Harv. JL \& Tech. 31
  (2017) 841.

\bibitem{dhurandhar2018explanations}
A.~Dhurandhar, P.-Y. Chen, R.~Luss, C.-C. Tu, P.~Ting, K.~Shanmugam, P.~Das,
  Explanations based on the missing: Towards contrastive explanations with
  pertinent negatives, arXiv preprint arXiv:1802.07623.

\bibitem{cui2015optimal}
Z.~Cui, W.~Chen, Y.~He, Y.~Chen, Optimal action extraction for random forests
  and boosted trees, in: Proceedings of the 21th ACM SIGKDD international
  conference on knowledge discovery and data mining, 2015, pp. 179--188.

\bibitem{kanamori2020dace}
K.~Kanamori, T.~Takagi, K.~Kobayashi, H.~Arimura, Dace: Distribution-aware
  counterfactual explanation by mixed-integer linear optimization, in:
  Proceedings of the Twenty-Ninth International Joint Conference on Artificial
  Intelligence, IJCAI-20, Christian Bessiere (Ed.). International Joint
  Conferences on Artificial Intelligence Organization, 2020, pp. 2855--2862.

\bibitem{grath_interpretable_2018}
R.~M. Grath, L.~Costabello, C.~L. Van, P.~Sweeney, F.~Kamiab, Z.~Shen,
  F.~Lecue, Interpretable {Credit} {Application} {Predictions} {With}
  {Counterfactual} {Explanations}, arXiv:1811.05245 [cs]ArXiv: 1811.05245.

\bibitem{van2019interpretable}
A.~Van~Looveren, J.~Klaise, Interpretable counterfactual explanations guided by
  prototypes, arXiv preprint arXiv:1907.02584.

\bibitem{pawelczyk2020learning}
M.~Pawelczyk, K.~Broelemann, G.~Kasneci, Learning model-agnostic counterfactual
  explanations for tabular data, in: Proceedings of The Web Conference 2020,
  2020, pp. 3126--3132.

\bibitem{laugel2018comparison}
T.~Laugel, M.-J. Lesot, C.~Marsala, X.~Renard, M.~Detyniecki, Comparison-based
  inverse classification for interpretability in machine learning, in:
  International Conference on Information Processing and Management of
  Uncertainty in Knowledge-Based Systems, Springer, 2018, pp. 100--111.

\bibitem{dandl2020multi}
S.~Dandl, C.~Molnar, M.~Binder, B.~Bischl, Multi-objective counterfactual
  explanations, arXiv preprint arXiv:2004.11165.

\bibitem{bliek1u2014solving}
C.~Bliek1{\'u}, P.~Bonami, A.~Lodi, Solving mixed-integer quadratic programming
  problems with ibm-cplex: a progress report, in: Proceedings of the
  twenty-sixth RAMP symposium, 2014, pp. 16--17.

\bibitem{artelt2020convex}
A.~Artelt, B.~Hammer, Convex density constraints for computing plausible
  counterfactual explanations, arXiv preprint arXiv:2002.04862.

\bibitem{russell2019efficient}
C.~Russell, Efficient search for diverse coherent explanations, in: Proceedings
  of the Conference on Fairness, Accountability, and Transparency, 2019, pp.
  20--28.

\bibitem{snell2017prototypical}
J.~Snell, K.~Swersky, R.~Zemel, Prototypical networks for few-shot learning,
  Advances in neural information processing systems 30.

\bibitem{jia2015new}
H.~Jia, Y.-m. Cheung, J.~Liu, A new distance metric for unsupervised learning
  of categorical data, IEEE transactions on neural networks and learning
  systems 27~(5) (2015) 1065--1079.

\bibitem{kaufman2009finding}
L.~Kaufman, P.~J. Rousseeuw, Finding groups in data: an introduction to cluster
  analysis, Vol. 344, John Wiley \& Sons, 2009.

\bibitem{van2019distance}
M.~van~de Velden, A.~Iodice~D'Enza, A.~Markos, Distance-based clustering of
  mixed data, Wiley Interdisciplinary Reviews: Computational Statistics 11~(3)
  (2019) e1456.

\bibitem{foss2019distance}
A.~H. Foss, M.~Markatou, B.~Ray, Distance metrics and clustering methods for
  mixed-type data, International Statistical Review 87~(1) (2019) 80--109.

\bibitem{article_causal}
J.~Pearl, Causal inference in statistics: An overview, Statistics Surveys 3
  (2009) 96--146.
\newblock \href {http://dx.doi.org/10.1214/09-SS057}
  {\path{doi:10.1214/09-SS057}}.

\bibitem{pearl2000models}
J.~Pearl, et~al., Models, reasoning and inference, Cambridge, UK:
  CambridgeUniversityPress 19.

\bibitem{shimizu2014lingam}
S.~Shimizu, Lingam: Non-gaussian methods for estimating causal structures,
  Behaviormetrika 41~(1) (2014) 65--98.

\bibitem{deb2000fast}
K.~Deb, S.~Agrawal, A.~Pratap, T.~Meyarivan, A fast elitist non-dominated
  sorting genetic algorithm for multi-objective optimization: Nsga-ii, in:
  International conference on parallel problem solving from nature, Springer,
  2000, pp. 849--858.

\bibitem{deb2002fast}
K.~Deb, A.~Pratap, S.~Agarwal, T.~Meyarivan, A fast and elitist multiobjective
  genetic algorithm: Nsga-ii, IEEE transactions on evolutionary computation
  6~(2) (2002) 182--197.

\bibitem{raquel2005effective}
C.~R. Raquel, P.~C. Naval~Jr, An effective use of crowding distance in
  multiobjective particle swarm optimization, in: Proceedings of the 7th annual
  conference on Genetic and evolutionary computation, 2005, pp. 257--264.

\bibitem{xu2017multi}
X.~Xu, Z.~Shi, Multi-objective based spectral unmixing for hyperspectral
  images, ISPRS Journal of Photogrammetry and Remote Sensing 124 (2017) 54--69.

\bibitem{whitley1994genetic}
D.~Whitley, A genetic algorithm tutorial, Statistics and computing 4~(2) (1994)
  65--85.

\bibitem{pymoo}
J.~{Blank}, K.~{Deb}, Pymoo: Multi-objective optimization in python, IEEE
  Access 8 (2020) 89497--89509.

\bibitem{magrini2017conditional}
A.~Magrini, S.~Di~Blasi, F.~M. Stefanini, A conditional linear gaussian network
  to assess the impact of several agronomic settings on the quality of tuscan
  sangiovese grapes, Biometrical Letters 54~(1) (2017) 25--42.

\bibitem{Dua:2019}
D.~Dua, C.~Graff, \href{http://archive.ics.uci.edu/ml}{{UCI} machine learning
  repository} (2017).
\newline\urlprefix\url{http://archive.ics.uci.edu/ml}

\bibitem{wightman1998lsac}
L.~F. Wightman, Lsac national longitudinal bar passage study. lsac research
  report series.

\bibitem{laugel2017inverse}
T.~Laugel, M.-J. Lesot, C.~Marsala, X.~Renard, M.~Detyniecki, Inverse
  classification for comparison-based interpretability in machine learning,
  arXiv preprint arXiv:1712.08443.

\bibitem{sharma2019certifai}
S.~Sharma, J.~Henderson, J.~Ghosh, Certifai: Counterfactual explanations for
  robustness, transparency, interpretability, and fairness of artificial
  intelligence models, arXiv preprint arXiv:1905.07857.

\bibitem{zheng2018feature}
A.~Zheng, A.~Casari, Feature engineering for machine learning: principles and
  techniques for data scientists, " O'Reilly Media, Inc.", 2018.

\bibitem{turner1999conceptual}
C.~R. Turner, A.~Fuggetta, L.~Lavazza, A.~L. Wolf, A conceptual basis for
  feature engineering, Journal of Systems and Software 49~(1) (1999) 3--15.

\bibitem{dong2018feature}
G.~Dong, H.~Liu, Feature engineering for machine learning and data analytics,
  CRC Press, 2018.

\end{thebibliography}

\end{document}